\documentclass[a4paper,conference]{IEEEtran}
\usepackage{amsmath,amsfonts,amssymb}
\usepackage{graphicx}
\usepackage{setspace}
\usepackage{tocloft}
\usepackage{multirow}
\usepackage{lineno}
\usepackage[table]{xcolor}
\usepackage{cite}
\usepackage{color}
\usepackage{comment}
\usepackage{amsmath,amssymb}
\usepackage[utf8]{inputenc}
\usepackage{booktabs}
\usepackage{floatrow}
\usepackage{subcaption}
\usepackage{algorithm}
\usepackage{algorithmic}
\usepackage{xhfill}

\ifCLASSINFOpdf
\else
\fi
\begin{document}

\title{A Multi-task Contextual Atrous Residual Network for Brain Tumor Detection \& Segmentation}

\author{\IEEEauthorblockN{Ngan Le, Kashu Yamazaki, Dat Truong}
\IEEEauthorblockA{University of Arkansas \\ Fayetteville, AR USA\\
Email: \{thile, kyamazak, tt032\}@uark.edu}
\and
\IEEEauthorblockN{Kha Gia Quach}
\IEEEauthorblockA{Concordia University \\ Montreal, QC, CANADA\\
Email: kquach@ieee.org }
\and
\IEEEauthorblockN{Marios Savvides}
\IEEEauthorblockA{Carnegie Mellon University \\ Pittsburgh, PA, USA\\
Email: msavvid@ri.cmu.edu}
}


\maketitle

\begin{abstract}
In recent years, deep neural networks have achieved state-of-the-art performance in a variety of recognition and segmentation tasks in medical imaging including brain tumor segmentation. We investigate that segmenting a brain tumor is facing to the \textit{imbalanced data problem} where the number of pixels belonging to the background class (non tumor pixel) is much larger than the number of pixels belonging to the foreground class (tumor pixel). To address this problem, we propose a \textit{multi-task network} which is formed as a cascaded structure. Our model consists of two targets, i.e., (i) effectively differentiate the brain tumor regions and (ii) estimate the brain tumor mask. The first objective is performed by our proposed \textit{contextual brain tumor detection network}, which plays a role of an attention gate and focuses on the region around brain tumor only while ignoring the far neighbor background which is less correlated to the tumor. Different from other existing object detection networks which process every pixel, our contextual brain tumor detection network only processes contextual regions around ground-truth instances and this strategy aims at producing meaningful regions proposals. The second objective is built upon a \textit{3D atrous residual network} and under an encode-decode network in order to effectively segment both large and small objects (brain tumor). Our 3D atrous residual network is designed with a skip connection to enables the gradient from the deep layers to be directly propagated to shallow layers, thus, features of different depths are preserved and used for refining each other. In order to incorporate larger contextual information from volume MRI data, our network utilizes the 3D atrous convolution with various kernel sizes, which enlarges the receptive field of filters. Our proposed network has been evaluated on various datasets including BRATS2015, BRATS2017 and BRATS2018 datasets with both validation set and testing set. Our performance has been benchmarked by both region-based metrics and surface-based metrics. We also have conducted comparisons against state-of-the-art approaches.\footnote{Code and models will be available upon the request}
    
\end{abstract}


\section{Introduction}
\label{sect:intro} 
With an estimation of 86,970 brain tumor patients in the United States (US) in 2019 and about 700,000 people in the US are living with a primary brain tumor. Brain tumor has become the highest per-patient initial cost of care for any cancer group. They can be deadly, significantly impact quality of life, and life-altering for a patient and their loved ones. 
Magnetic resonance imaging (MRI) and computed tomography (CT) scans are two popular imaging tests for brain or spinal cord tumor diagnosis. Different from CT scans which focus on skulls, MRI scans are very good for looking at the brain and spinal cord, which are considered the best way to look for tumors in these areas. Current recommendations for standardized brain tumor MRI contrast modes are: Fluid attenuation inversion recovery (FLAIR) which pronounces whole tumor, spin-lattice relaxation T1-weighted (T1), pin-spin relaxation T2-weighted (T2) which pronounces tumor core, T1-weighted contrast-enhanced (T1c) which pronounces enhancing tumor structures. From large amounts of MRI images generated during clinical routine, manually detecting, segmenting brain tumors for cancer diagnosis is costly and time consuming. In addition to being diffused, poorly contrasted, and extended tentacle-like structures, brain tumors can appear anywhere in the brain and vary in their shape, size and structure. Furthermore, tumor mass effect \cite{Menze_brats2015} caused by the variety of appearance of surrounding normal tissues  also makes the problem  of brain tumor segmentation become more challenging. Thus, the task of automatically segmenting brain tumor is very challenging. 

\textbf {Our proposed multi-task network for  automatically detecting \& segmenting brain tumor is based on the following observations:}
\begin{itemize}
    \item Brain tumor regions only occupy small parts of the brain image, thus, it is computationally expensive and unproductive if segmentation task is preformed at every single pixel over the entire brain image because the segmentation task is a pixel-level processing. 
    \item Brain tumor segmentation is considered as a binary classification at pixel level where each pixel is classified as either tumor pixel (foreground) or not tumor pixel (background). Under this problem setting up, the number of pixels belonging to foreground class is quite small compared to the number of pixels belonging to background class. This imbalanced data problem makes the requirement of high accurately segmenting brain tumor become more difficult.
    \item Brain tumor contains three components i.e. Whole tumor (WT), Tumor Core (TC), Enhancing Tumor (ET).  These components are nested and structured with strong correlation.
    \item Most existing deep networks cannot directly propagate the gradient information from deep layers to shallow layers because of the convolutional layer between features of different resolutions.
    \item The existing networks  \cite{Ronneberger_2015, 3DUnet, Dong_2017, Wang_TTA, Kamnitsas_2018, Le2018_3} have been successfully applied to medical imaging but may face to the vanishing problem once going too deep.
\end{itemize}

\textbf{Our contribution can be summarized as follows:}
\begin{itemize}
    \item Propose a \textbf{multi-task network} where the first task aims at localizing the region of interest in order to \textbf{solve the class imbalanced data problem} by eliminating redundant background pixels which are less correlation with brain tumor region. 
    \item Propose a \textbf{contextual detection network} which only processes contextual regions around the ground-truth of brain tumor instances instead of processing every single pixel in an image pyramid. The proposed contextual detection network helps to effectively generate more meaningful region proposals with less time consuming. Our contextual detection network plays a role of an attention gate that pays attention to the brain tumor region and its near neighbors whereas its far neighbors like black background will be ignored.
    \item To learn larger contextual information, our network makes use of \textbf{3D atrous convolution}, which controls the field-of-view and finds the best trade-off between accurate localization and context assimilation.
    \item In order to extract higher feature with deeper network, our network is designed with \textbf{skip connections} under a residual network to avoid the vanishing problem as well as enable the gradient from the very deep layers to be \textbf{directly propagated to shallow layers}.
\end{itemize}

\section{Related Work}

Brain tumor segmentation is the division and classification of brain tumor tissue into four tumor components: edema (whole tumor), non-enhancing solid core (tumor core), necrotic/cystic core, and enhancing core. 

Machine learning and image processing have been used to solve the problem of brain tumor segmentation. Support Vector Machines was effectively used to segment brain tumor in \cite{Havaei_2016} whereas  \cite{Bauer_2011} incorporates Conditional random fields to refine the segmentation.  K-means is also a productive method to address this issue such as in \cite{Anitha_2016}. Among many image processing approaches, Level Sets(LS) is one of the most classic methods to segment brain tumor as in \cite{Taheri_2010, Sean_2002}. 

In the last couple years, we have seen great deal of deep learning which has become one of the most productive methods in many areas \cite{He2016, Le2017_1, Le2018_1, Le2018_2, Simonyan2015},  
including segmenting brain tumor \cite{Havaei_2015, Ronneberger_2015, Pavel_2015, 3DUnet, Pereira_2016, Kamnitsas_2016_DeepMedic, Kamnitsas_2017, Le2018_3}. Havaei, et al. \cite{Havaei_2015} proposed one of the first deep network architecture for segmenting brain tumors in the BRATS 2015 challenging. In order to improve the performance and overcome the limitation of training data, Convolutional Neural Networks (CNNs) are designed in a fashion which combines with other classification methods or clustering methods \cite{Pavel_2015}. Later, Pereira, et al. \cite{Pereira_2016} evaluated a 11-layered CNN architecture on BRATS dataset by implementing small 3 x 3 sized filters in the convolutional layers, thereby reducing the total number of network parameters. Kamnitsas, et al. \cite{Kamnitsas_2017} were the first to propose a patch-based 3D CNN for brain tumor segmentation. Instead of processing MRI scans slice-by-slice as in previous 2D methods, 3D patches are extracted from each MRI channel and fed through four 3D convolutional layers. Recently, Le, et al. \cite{Le2018_3} made use of Gated Recurrent Unit to reform Level Set curve evolution to successfully segment brain tumor. Their network is built upon Fully Convolutional Network (FCN) under recurrent network defined by Level Set curve evolution. Cascaded network is classic yet powerful architecture network that has boosted performance on various tasks and Wang, et al. \cite{Guotai_Cascade_2017} proposed a cascaded anisotropic convolutional neural networks where the whole tumor is segmented in the first step and the bounding box of the result is used for the tumor core segmentation in the second step. The enhancing tumor core is then segmented based on the bounding box of the tumor core segmentation result. Later, Wang, et al. \cite{Wang_TTA} has demonstrated the usefulness of using test-time augmentation which help to improve CNNs' performance for brain tumor segmentation. In their proposed method, they also use different underpinning network structures and augmented the image by 3D rotation, flipping, scaling and adding random noise at both training and test time. In order to robustly perform through aggregation of predictions from a wide range of methods, Kamnitsas, et al. \cite{Kamnitsas_2018} made use of Ensembles of Multiple Models and Architectures (EMMA). This approach  aims at reducing the influence of the meta-parameters of individual models and the risk of overfitting the configuration to a particular database.

One of the most successful deep network in medical imaging is Unet proposed by Ronneberger, et al. \cite{Ronneberger_2015}. Unet consists of a downsampling FCN followed by an upsampling FCN known as the network's contractive and expansive paths. Because of the U-Net's success on BRATS, many other brain tumor segmentation methods such as \cite{Dong_2017} and \cite{3DUnet} further modified the U-Net architecture for 3D convolution. 3D CNNs with residual connections were also proposed in Deep Medic \cite{Kamnitsas_2017} which is an another successful deep learning approach in brain tumor segmentation. These 3D U-nets were shown to outperform current 2D medical imaging segmentation models in prostate and kidney segmentation, respectively. 

\section{Proposed Network}
Our proposed network is designed as a cascade structure with two targets, namely, (i) reduce effect of the imbalanced data problem by localizing the brain tumor region and removing redundant background pixels which are far from the tumor region and less correlated to the brain tumor region; (2) effectively extract brain tumor pixels out of the background and classify the tumor pixels as either enhancing tumor (ET) or tumor core (TC) or whole tumor (WT). The first task is performed through our proposed \textbf{Contextual Detection Network}(in section \ref{sec:detection}) whereas the second task is done by our \textbf{3D Atrous Residual Network} (in section \ref{sec:segmentation}).

\subsection{Contextual Detection Network}
\label{sec:detection}
Our proposed contextual detection network is an extension of SNIPER \cite{sniper}, which is based on Faster-RCNN implementation. Therefore, we first revise the well-known region-based object detection networks including R-CNN, Fast-RCNN and Faster-RCNN in section \ref{sec:revise}. We then detail our proposed contextual detection network for detecting brain tumor region in section \ref{sec:proposed}.

\subsubsection{Object Detection Networks: A revise}
\label{sec:revise}
One of the most important approaches to the object detection and object classification problems is Region-based CNNs
\cite{Uijlings13_rcnn}, \cite{Girshick2015}, \cite{girshick2015fast},\cite{faster_rcnn}. 
R-CNN detector \cite{Uijlings13_rcnn} is considered one of the primary deep learning based object detection algorithms. In  general, R-CNN contains three steps: (i) generate a set of object proposals; (ii) resize these proposals to fixed size; (iii) compute object score for each  proposal. 
This model is scale invariant, but the computational cost for training and inference is high. To address the computational bottleneck, Girshick Ross \cite{girshick2015fast} proposed Fast-RCNN to project region proposals to a high level convolutional feature map and use the pooled features as a semantic representation for region proposals. In this approach, the convolutional layers are shared and only lightweight fully connected layers are applied on each proposal. Furthermore, to deal with objects of different sizes, Fast-RCNN \cite{girshick2015fast} performs an inference process at multiple scales and the detecting results from multiple scales are combined. Different from R-CNN where all proposal is resize to a fixed size regardless large object or small object, Fast-RCNN upsamples and downsamples every proposal in the image in multiple scales. 
The Fast-RCNN is not only computationally effective thanks to sharing the convolutional features, but also better capture more contextual information at different scales. It has become a well-known approach for object detection. 
Fast R-CNN accelerates the detection network using the ROI-pooling layer. 
However the region proposal step is designed out of the network hence still remains a bottleneck.

Faster R-CNN \cite{faster_rcnn} addresses this problem by introducing the Region Proposal Network (RPN). A RPN is implemented in a fully convolutional style to predict the object bounding boxes and the object probability scores. In addition, the anchors are defined with different scales and ratios to achieve the translation invariance. The RPN shares the full-image convolution features with the detection network. 

Based on the aforementioned observations that brain tumor regions occupy only a small part of the brain image, our target is to process meaningful contextual regions instead of working on every pixel of the whole image. In general, our detector is an extension of Faster-RCNN \cite{faster_rcnn} together inherits the merits of both R-CNN \cite{Uijlings13_rcnn} and Fast-RCNN \cite{girshick2015fast} by generating scale specific context-regions that cover maximum proposals at a particular scale. Like Fast-RCNN, our detector classifies all the proposals inside context-regions. Different from Fast-RCNN which does not care about proposal size and performs upsampling and downsampling at every proposal, our detector does not upsample the image where there are large objects and ignore all easy background regions like black region surrounding brain for example. By doing those, our detector is significantly faster than Fast-RCNN detector trained on an image pyramid.

\subsubsection{Contextual Detection Network}
\label{sec:proposed}
Working in similar fashion to Faster-RCNN, our proposed network assign each proposal generated by RPN five values corresponding to object score, bounding box position ($x$, $y$, $w$, $h$). However, different from Faster-RCNN which processes every pixel in a given image, our detector targets to obtain more "meaningful" proposals as well as reduce time consumption during training procedure by employing SNIPER detector \cite{sniper}. There are two components in the contextual detection network, namely, proposals generation and label assignment as follows:

\begin{figure}
\centering
\includegraphics[width=9cm]{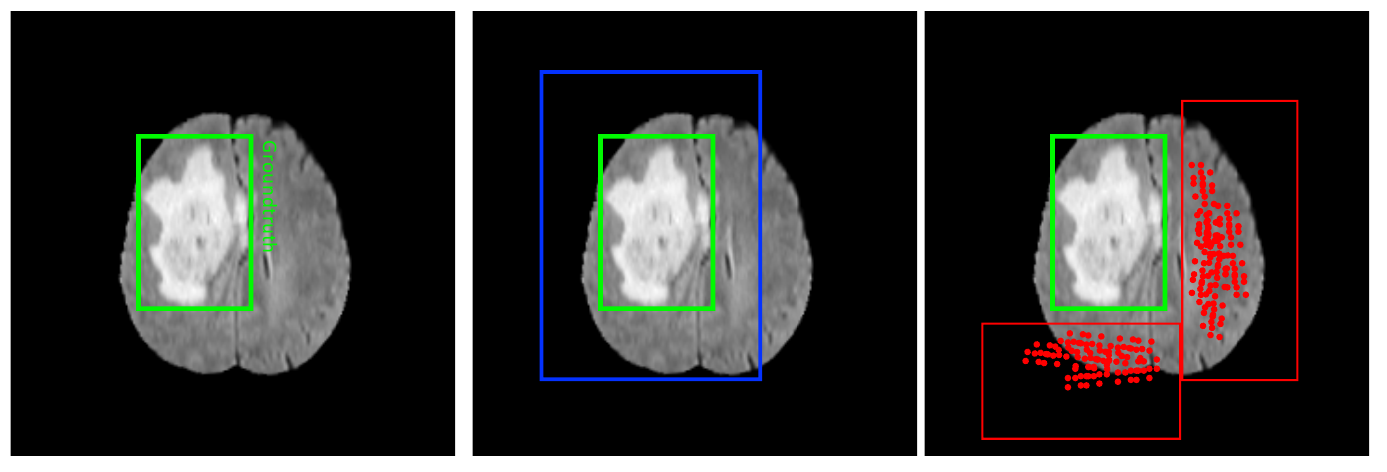}
\caption{Illustration of contextual detection. From left to right: Given an image with \textcolor{green}{groundtruth} (left), \textcolor{blue}{positive window} is defined as contextual around groundtruth (middle) and \textcolor{red}{negative windows} which are not covered in the positive windows (right). Each red circle is corresponding to center of each proposal.}
\label{fig:detector}
\end{figure}

\begin{figure*}
\centering
\includegraphics[width=17cm]{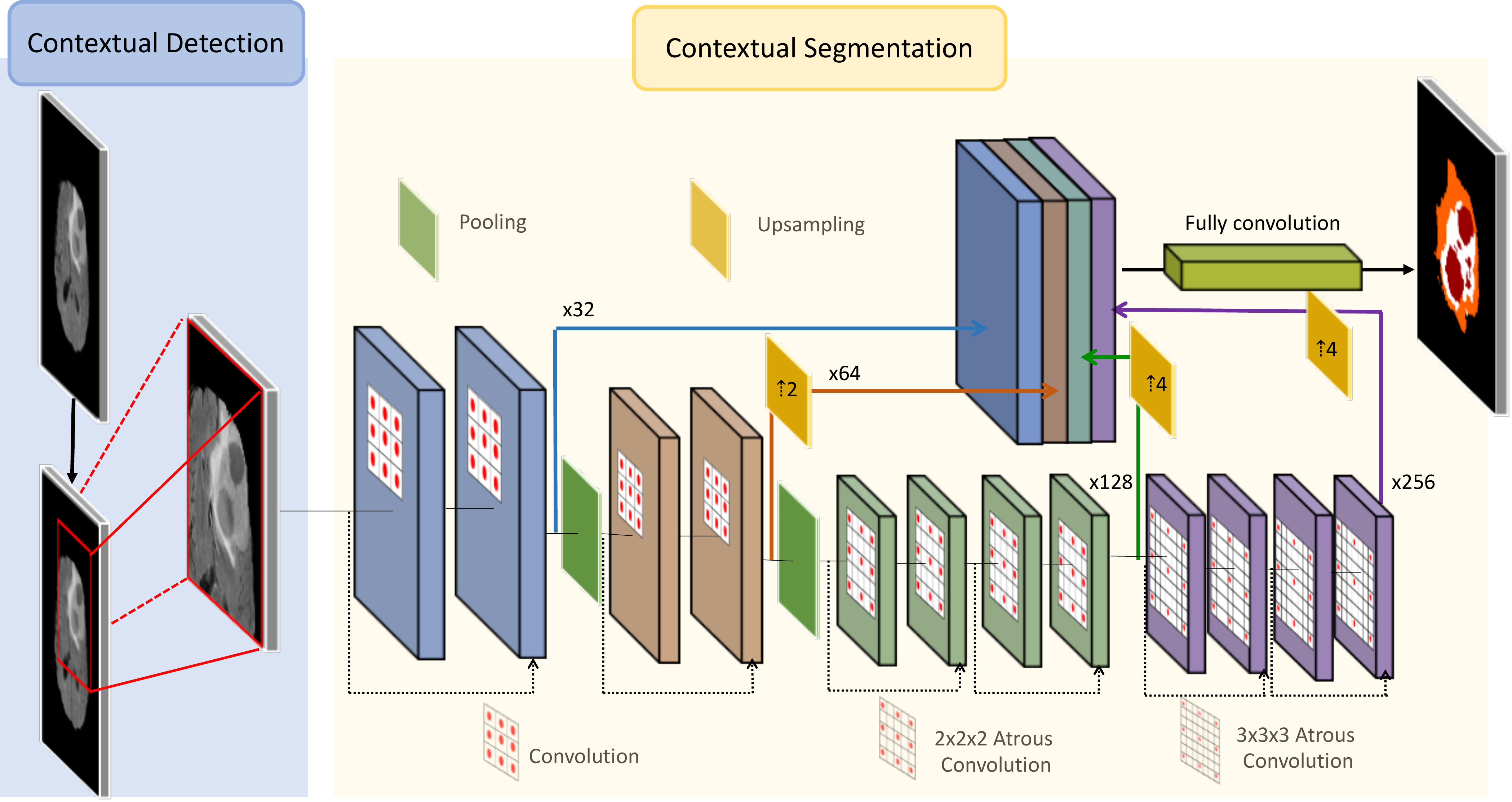}
\caption{The architecture of our proposed multi-task contextual atrous residual  network. There are 2 atrous convolutions i.e. $2 \times 2 \times 2$,  $3 \times 3 \times 3$, are used. In this network, $-----$ represents a residual connection.}
\label{fig:network_flowchart}
\end{figure*}
 
In order to generate positive proposals and negative proposals, we first narrow down the searching space based on groundtruth instances i.e. the brain tumor. In our detection network, we focus on the brain tumor region and its near neighbors instead of far neighbors which are less correlated to the brain tumor or brain. At each scale $s_i$, a given image is first resized to $w_i \times h_i$, then partitioned into $k_1 \times k_2$ pixel region windows with the intervals of $K = 4$ pixels. As a result, we obtain a two-dimensional array of windows at each scale. In our contextual detection network, the \textit{positive proposals are assumed to be in a positive windows and negative proposals are assumed to be in a negative windows}. 

\textbf{Positive windows generation}:  Different from $p$ positive chips selection in SNIPER \cite{sniper}, groundtruth proposals in our detection network are created by union regions of whole tumor (WT), tumor core (TC) and enhance tumor (ET). Thus, the positive window is defined as a contextual region around the groundtruth and the size of the positive window is as twice as brain tumor groundtruth. An example of positive windows generation is given in Fig.\ref{fig:detector}(middle) where the groundtruth is presented in green box and the positive window is presented in blue box.

\textbf{Negative Windows Generation}: Although the positive windows cover all the positive proposals, a significant portion of the brain, which is not lesion and maybe considered to be background, is not covered by them. Compare to the lesion region, the background including brain background and black background is much bigger. In order to avoid high computation, a simple approach is to employ object proposals to identify regions where objects are likely to be present. In this step, we eliminate regions that does not contains any object, simply the almost black background will be ignored. Notably, object in this scenario is brain background but not brain tumor. In order to obtains negative windows, we first train a RPN which is setup as same as Faster-RCNN \cite{faster_rcnn} for small number of epoches over the entire training set to roughly guide us in selecting regions which are likely to contain false positives. If no proposals are generated in a major portion of the image by RPN, then it is unlikely to contain an object instance. At each scale, to generate negative windows, we first eliminate all the proposals which have been covered in the positive windows in the previous step, then greedily select all the windows which cover at least $P$ proposals ($P$ is chose as 2 in our experiments). An example of negative windows generation is given in Fig.\ref{fig:detector}(right) where the groundtruth is presented in green box and the negative window is presented in red box which are not covered in the positive windows. Each red circle is corresponding to center of each proposal.

\textbf{Training Detection Network}: Our contextual detection network is trained on aforementioned positive/negative windows and based on the implementation of Faster-RCNN. During training, we ignore proposals, which do not fall in either positive windows or negative windows. That helps to remove a large of number redundant proposals and just keep a small number of valid proposals. Like Faster-RCNN, any proposal which has an overlap greater than 0.5 with a ground-truth box is marked as positive and assign bounding-box targets for the proposal. In experiment, we generate 300 proposals for each positive/negative window. For classification, we use simple softmax cross-entropy loss. 

\subsection{3D Atrous Residual Segmentation Network}
\label{sec:segmentation}

This section describes how we design the proposed 3D atrous residual network. The segmentation network takes the detected result from our previous detector component which is first extended on each direction an offset $f= 6$ and then resized to $W \times H$ as the input. In the experiment, we choose $W = H = 64$ as the width and the height of the input. Let $W \times H \times C \times N$ as the input of segmentation network, where $C$ is the number of modalities and $C = 4$ corresponding to Flair, T1, T1ce, T2 and $N$ is the batch size and we choose $N = 4$ in our experiment. The entire proposed network with two sub-networks corresponding to detection and segmentation is given in Fig.\ref{fig:network_flowchart}. There are four main basic components in the segmentation network including vanilla convolution layers, atrous convolution layers (a.k.a dilated convolution), pooling layers and feature map concatenation. Because the input of the network is 3D volumetric data, all the convolution filters are 3D convolution filters. Particularly, in the proposed segmentation network, the highest resolution and half resolution layers are designed with vanilla convolution whereas the quarter resolution layers are designed with $2\times2$ and  $3\times3$ atrous convolutions in order to learn long-range representation. In the proposed deep learning framework,  the vanilla convolutions with small receptive field are used in the shallower layers to learn local feature whereasthe  atrous convolutions with larger receptive field are used in the deeper layers to learn longer range contextual information (global feature) without the need for pooling. Atrous convolution layers have been shown to decrease blurring in semantic segmentation maps \cite{Dialatied_Conv, blurring_seg}.

The final feature map is a combination of intermediate feature maps at different resolution. The depth of the feature maps (at each resolution) keeps increasing (32, 64, 128, 256) as shown in Fig.\ref{fig:network_flowchart} where the channel size of the final feature map is $= 32 + 64 + 128 + 256 = 480$). 

\begin{figure}
\centering
\includegraphics[width=9cm]{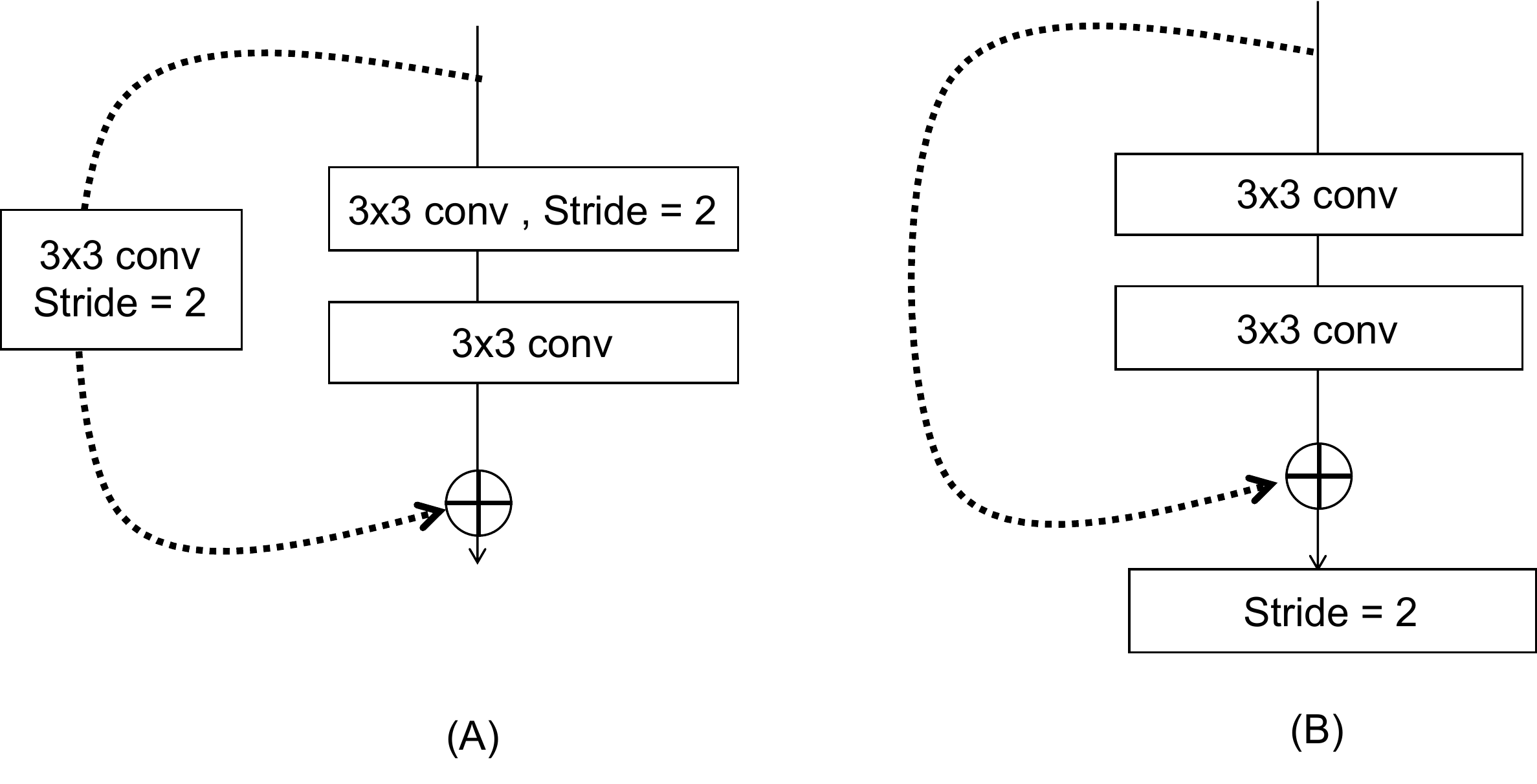}
\caption{The original ResNet block architecture (A), The proposed resnet block (B)}
\label{fig:resnet_weakness}
\end{figure}
 
 In computer vision, image segmentation is known as pixel-level tasks which require large feature maps to keep the resolution. Up-sampling strategy has been successfully applied to address the issue of high resolution in the pixel-level task. The series of works including U-Net\cite{Ronneberger_2015} CRLS \cite{Le2018_2}, CRRN \cite{Le2018_1}, DRLS \cite{Le2018_3}, 3D-Unet \cite{Kamnitsas_2017} have all shown their capability in pixel-level tasks. However, such network architectures still use convolution between features of different scales (resolutions), which cannot preserve the representations. Besides, they do not provide an up-sampling pathway to enable features with large resolution and more semantic meaning. In order to communicate between different resolution, features from different depth are usually combined by concatenate \cite{Hariharan2015_Hypercolumn}, residual connection \cite{He2016}. However, features from deep layers and shallow layers were not used for refining each other in concatenating mechanism. Whereas residual blocks with addition has problem when directly propagated from the very deep layers to shallow layers. This problem is caused by the convolutional layer between features of different resolutions. As shown in Fig.\ref{fig:resnet_weakness} (A), the skip connection in Resnet \cite{He2016} contains a convolutional layer with stride 2 to deal with the inconsistency between the numbers of input channels and output channels. In our proposed network, we concatenate features of very different depths to the final output as shown in Fig.\ref{fig:resnet_weakness} (B). By concatenating features from different scales, the semantic meaning of features are also preserved throughout the whole network.

For the loss function, we make use of Dice coefficient which is essentially a measurement of overlap between two samples. This measure ranges from 0 to 1 where a dice coefficient of 1 denotes perfect and complete overlap. The dice loss can be calculated as:
\begin{equation}
    \mathcal{L}_{(P, Y)} = 1 -  \frac{2|P \cap Y|}{|P| + |Y|}
\end{equation}
where $\frac{2|P \cap Y|}{|P| + |Y|}$ is known as a dice coefficient. $2|P \cap Y|$ represents the common elements between predicted mask $P$ and groundtruth mask $Y$. In our proposed method, we imperially make use of soft dice loss which directly computes on the predicted probabilities instead of thresholding and converting them into a binary mask.

\begin{table*}[h]
    \centering
    \caption{The evaluation (\%) of our proposed network with both area-based evaluation and surface-based evaluation on online testing set of BRATS 2018, BRATS 2017, BRATS 2015 with all area-based metrics and surface-based metrics. WT: Whole Tumor, TC: Tumor Core, ET: Enhancing Tumor.}
    \label{Tab:Our_performance_metrics}
    \vspace{10mm}
    \begin{tabular}{|c|c|ccc|ccc|ccc|}
    \hline
    &  Metrics & \multicolumn{3}{c|}{BRATS 2018} & \multicolumn{3}{c|}{BRATS 2017} & \multicolumn{3}{c|}{BRATS 2015} \\ 
     
     &  & WT       & TC       & ET       & WT       & TC       & ET       & WT       & TC       & ET       \\ \hline 
     \multirow{4}{*}{Area based} 
     
    & {Dice score} & 86.4     & 82.5     & 78.2     & 85.9     & 82.2     & 74.2     & 90.4     & 82.3     & 72.4     \\ 
    & {Sensitivity-TPVF} & 95.4     & 88.0     & 82.8     & 95.8     & 85.8     & 79.3     & 92.1     & 83.4     & 76.2     \\ 
    & {Specificity-TNVF} & 98.7     & 99.6     & 99.8     & 98.5     & 99.6     & 99.8     & 91.3     & 82.2     & 73.1     \\ 
   & {Precision}&     87.8     &     86.9     &  84.9        &    86.7      &    85.3      &     78.3     &    88.4      &   86.5       &   74.3       \\ \hline 
    \multirow{2}{*}{Surface based}
    & {Hauf}  &    4.7      &     6.3    &    5.2      &    8.8      &    10.1      &     11.2     &  5.1         &  12.8        &      6.8    \\ 
    & {ASSD}  & 1.03 &   1.93       &   2.48       &     1.06     &   2.23       &     2.94     &     0.95     &    2.35      &     3.46      \\ \hline 
    \end{tabular}
 
\end{table*}

\section{Experiments}
\subsection{Datasets}
We evaluate the proposed method on three most recent publicly available brain tumor datasets, i.e., BRATS 2015, BRATS 2017 and BRATS 2018. The BRATS 2018 database includes 68 validation MRI scans and 285 3D training MRI scans, which are composed of 210 HGG scans and 75 LGG scans. The BRATS 2017 database includes 47 validation MRI scans and 285 3D MRI scans, which are simmilar the training set in BRATS 2018. The BRATS 2015 database includes 110 validation MRI scans and 274 training MRI scans which as composed of  220 HGG scans and 54 LGG scans. For each scan, there are 4 available modalities, i.e., T1, T1C, T2, and Flair. Each image is registered to a common space, sampled to an isotropic $1mm \times 1mm \times 1mm$ resolution by the organizers and has a dimension of 240 x 240 x 155 \cite{6975210}. In the BRATS 2015 database, the tumor tissues are divided into four classes, i.e., necrotic, edema, non-enhancing tumor, and enhancing tumor. In the BRATS 2017, BRATS 2018 databases, the necrotic and non-enhancing tumors are grouped into one category; therefore, there are only three tumor classes in this database. To evaluate the effectiveness of the proposed methods, we setting up the following experiments on both validation and testing sets, namely, (i) BRATS 2018 online testing set; (ii) BRATS 2017 online testing set; (iii) BRATS 2015 online testing set; BRATS 2017/2018 testing set. For the experiments on validation set, we randomly select 67 scans (42 HGG scans and 15 LGG scans) from all the 285 cases in the BRATS2017/BRATS 2018 database to construct a local validation subset. The remaining scans are utilized as the training set.

\begin{table}[]
\centering
\caption{Performance of the \textbf{proposed 3D Atrous Residual Network with Contextual Detection } and comparison to state-of-the-art methods on BRATS 2018 \textbf{validation set} with Dice score and Hausdorff (Hauf) distance. WT: Whole Tumor, TC: Tumor Core, ET: Enhancing Tumor. }
\begin{tabular}{l|lll|lll|}
\cline{2-7}
& \multicolumn{3}{c|}{Dice} & \multicolumn{3}{c|}{Hauf} \\ \cline{2-7} 
& WT      & \quad TC     & \quad ET     & WT      & \quad TC     & \quad ET     \\ \hline 
\multicolumn{1}{|l|}{\cite{3DUnet}}  & 86.38   & 76.58  & 73.44  & 12.00   & 10.37   & 9.37        \\ \hline 
\multicolumn{1}{|l|}{\cite{Guotai_Cascade_2017}}    & 90.05   & 83.78   & 78.59 & 3.89    & 6.47    & 3.28        \\ \hline 
\multicolumn{1}{|l|}{ \cite{Le2018_3}}    & 85.87   & \textbf{88.61}  & 77.08  & -       & -       & -       \\ \hline 
\multicolumn{1}{|l|}{\cite{Kamnitsas_2018}} & 90.1 & 79.7 & 73.8 & 4.23 & 6.56 & 4.50 \\ \hline 
\multicolumn{1}{|l|}{\cite{Wang_TTA}}   &    \textbf{90.50} & 83.80 & 78.60  & \textbf{3.89} & 6.48 & 3.28\\ \hline 
\multicolumn{1}{|l|}{\cite{Brats_2018_compare2}}  & 90.44 & 84.94 & \textbf{80.52} & 6.38 & \textbf{6.37} & \textbf{2.78}\\ \hline 
\multicolumn{1}{|l|}{\cite{3DUnet} } & 88.76 & 82.57 & 78.09 & 12.63 & 13.33 & 7.74 \\ \hline 
\multicolumn{1}{|l|}{\textbf{Ours}}    & \textbf{90.95}   & \textbf{88.88}  & \textbf{81.41}  & \textbf{3.73 }    &\textbf{ 5.92}     &\textbf{ 2.70}      \\ \hline 
\end{tabular}
\label{Tab:2018_testing}
\end{table}

\subsection{Evaluation Metrics}
The simplest evaluation metrics are True Positive, True Negative, False Positive and False Negative. 
However, it is hard to infer percentage of overlap between predicted segmentation and ground truth images from the above metrics. In this work, we evaluate the performance of the proposed method using the following standard metrics as suggested in BRATS challenge \cite{6975210},  \cite{Varduhi_2019}. The following metrics focus on both area-based evaluation and surface-based evaluation. The first evaluation, which makes use of area-based metrics, compares the difference between the area of segmentation and the groundtruth area. The second evaluation, which makes use of surface-based metrics, compares the difference between the contour of segmentation results and the groundtruth 

\subsubsection{Area-based metrics}
The similarity between predicted segmentation results and manual segmentation results (groundtruth) can be assessed using overlap measures as follows:
\begin{itemize}
\item \textbf{Dice Score (DSC):} The algorithm generates a predictions $P$ which is the segmentation of a tumor region from a modality. $P$ $\in\{0,1\}$ for each of the three tumor regions. The corresponding experts' consensus truth $T \in\{0,1\}$ is obtained from ground truth images for each of the regions. Evaluation metric Dice score is calculated as:
$ DSC(P,T) = \frac{2 \times |P_1 \wedge T_1|}{(|P_1| + |T_1|)} $, where $\wedge$ is the logical AND operator, $| \ |$ is the size of the set (i.e., the number of voxels belonging to it), and $P_1$ and $T_1$ represent the set of voxels where P=1 and T=1, respectively. The Dice score normalizes the number of true positives to the average size of the two segmented areas. It is identical to the F score (the harmonic mean of the precision recall curve) and can be transformed monotonously to the Jaccard score. \\
\item \textbf{Sensitivity(Sens): } or the true positive rate is the value representing the percentage of recognition of actual value. This metrics is equivalent to Recall as well as is know is true positive (TPVF), $Sens(P,T)  = \frac{|P_1 \wedge T_1|}{|T_1|}$.\\
\item \textbf{Specificity(Spec): } or true negative rate is the value representing the percentage of recognition of actual negatives. Accuracy is the degree of closeness of measurements of a quantity to its actual (true) value. Specificity score is also known as true negative (TNVF), $Spec(P,T) =  \frac{|P_0 \wedge T_0|}{|T_0|}$, where $P_0$ and $T_0$ represent voxels where $P=0$ and $T=0$, respectively. \\
\item \textbf{Precision (Pre): } Precision normalizes the volume of the correctly segmented region over the volume of the result of the segmentation,  $Pre(P,T) = \frac{|P_1 \wedge T_1|}{|P_1|}$.

\end{itemize}

Dice score, Sensitivity (TPVF), Specificity(TNVF), and Precision are measures of voxel-wise overlap of the segmented regions which are also known as area-based metrics.

\begin{table*}[h]
	\centering
	\caption{Comparison of our proposed \textbf{3D Atrous Residual Network} against other methods on BRATS 2015\textbf{ online testing sets} with Dice and Sensitivity metrics. WT: Whole Tumor, TC: Tumor Core, ET: Enhancing Tumor.} 
	\label{Tab:2015_online_testing_compare}
	\vspace{10mm}
	\begin{tabular}{|c|ccc|ccc|}
		\hline
		Methods                        & \multicolumn{3}{c|}{Dice Score} & \multicolumn{3}{c|}{Sensitivity} \\ 
		& WT        & TC       & ET       & WT        & TC        & ET       \\ \hline 
		Pereira et al\cite{Pereira_2016}          & 78.0      & 65.0     & 70.0      & -         & -         & -\\ \hline 
		Pavel et al\cite{Pavel_2015}             & 83.0      & 75.0     & 77.0     & -         & -         & - \\ \hline 
		Chang et al\cite{Peter_2016}            & 87.0      & 81.0    & 72.0    & -         & -         & -       \\ \hline 
		Deep Medic\cite{Kamnitsas_2016_DeepMedic} & 89.6     & 75.4    & 71.8    & 90.3     & 73.0     & 73.0  \\ \hline 
		DMRes\cite{Kamnitsas_2017}                  & \textbf{89.8}     & 75.0   & 72.0   & 89.1     & 72.1     & 72 .5  \\ \hline 
		Improved Unet \cite{Fabian_2017} & 85.0 & 74.0 & 64.0 & \textbf{91.0}& 73.0 & 72.0\\ \hline 
		DRLS\cite{Le2018_3}                & 88.0      & \textbf{82.0 }   & \textbf{73.0}    & \textbf{91.0}     & 76.0      & \textbf{78.0} \\ \hline 
		FSENet\cite{Chen_FSNet_2018}                & 85.0      & 72.0    & 61.0     & 86.0      & 68.0    & 63.0  \\ \hline 
		
		Multi\_task \cite{Chenhong_2019} & 87.0 & 75.0 & 65.0 & 89.0 & \textbf{85.0} & 63.0\\ \hline 
		\textbf{Our }             & \textbf{90.4}      & \textbf{82.3}     & \textbf{72.4}     & \textbf{92.0}     & \textbf{83.4}   & \textbf{76.7}   \\ \hline 
	\end{tabular}
\end{table*}

\begin{table*}[]
\centering
\caption{Comparison between with and without contextual detection network on BRATS 2018 validation set: An \textbf{ablation} to study the contribution of detection network. WT: Whole Tumor, TC: Tumor Core, ET: Enhancing Tumor.}
\label{tab:ablation}
\begin{tabular}{|c|c|c|c|c|c|c|}
\hline
\multirow{2}{*}{}                                                                             & \multicolumn{3}{c|}{Dice} & \multicolumn{3}{c|}{Hauf} \\ \cline{2-7} 
                                                                                              & WT      & TC     & ET     & WT      & TC     & ET     \\ \hline
\begin{tabular}[c]{@{}c@{}}Without Detection, \\ Only 3D Atrous Residual Network\end{tabular}   & 90.25   & 85.78  & 79.59  & 3.81    & 6.14   & 2.98   \\ \hline
\begin{tabular}[c]{@{}c@{}}Contextual Detection \& \\ 3D Atrous  Residual Network\end{tabular} & 90.95   & 88.88  & 81.40  & 3.73    & 5.92   & 2.70   \\ \hline
\end{tabular}
\end{table*}

\subsubsection{Surface-based metrics}
There are two common surface metrics, namely, Hausdorff distance and ASSD (Average symmetric surface distance) as follows:
\begin{itemize}
    \item \textbf{Hausdorff distance (Hauf)} is an evaluation metric that calculates the distance between segmentation boundaries, i.e., the surface distance. As shown by \cite{6975210} Hausdorff distance calculated for all points p on the surface $\delta P_1$ of a given volume $P_1$ the shortest least-squares distance d(p,t) to points t on the surface $\delta T_1$ of the other given volume $T_1$, and vice versa, finally returning the maximum value over all d as defined below:
\begin{equation}
    \small
{\rm Hauf}(P,T) = \max\{\sup_{p \in \partial P_1} \, \inf_{t \in \partial T_1} d(p,t),\, \sup_{t \in \partial T_1} \, \inf_{p \in \partial P_1} d(t,p)\}
\end{equation}
Hausdorff measure is highly sensitive to small outlying subregions(outliers) because it returns maximum over 'all' surface distances. 

\item \textbf{ASSD (Average symmetric surface distance)}: ASSD is calculated using surface voxels from two segmentations $P$ and groundtruth $T$. For each surface voxel from segmentation $P$, the Euclidean distance to the closest surface voxel of $T$ is calculated. The ASSD is the average of all distances calculated from $P$ to $T$ i.e. $d(s_p, S^T)$ and from $T$ to $P$ i.e. $d(s_t, S^P)$, where $s_p$, $s_t$ are points on the surface and $S^P$, $S^T$ are the total surface voxels in $P$ and $T$, repsectively.  
\begin{equation}
\begin{split}
    ASSD = \frac{\left( \sum_{s_p \in S^P} d(s_p, S^T) + \sum_{s_t \in S^T} d(s_t, S^P) \right)}{|S^P| + |S^T|}
\end{split}
\end{equation}

\end{itemize}

\subsection{Experimental Results}
Our proposed 3D atrous residual network with contextual detection is designed under a cascade structure and is evaluated on both validation set and testing set of BRATS 2015, BRATS 2017, and BRATS 2018. There are four different experiments conducted in this section: (i) BRATS 2017/BRATS 2018 validation set; (ii) BRATS 2017 online testing set; (iii) BRATS 2018  online vatestinglidation set; (iv) BRATS 2015  online testing set. For each experiment, we consider three types of tumors: whole tumor (WT), tumor core (TC), enhanced tumor (ET) which are extracted from dataset at different modalities, namely Flair, T$_2$, T$_1c$ respectively.

Table \ref{Tab:Our_performance_metrics} summarizes the performance of our proposed network on three different online testing sets, i.e. BRATS 2018, BRATS 2017, and BRATS 2015 on both area-based metrics and surface-based metrics. Table \ref{Tab:2018_testing} provides the comparison between our performance against state-of-the-art on BRATS2017/BRATS2018 validation set. In this table, the best and second best scores are highighted. Compared to the state-of-the-art, our proposed network achieve the best score in most of the metrics. The comparison on BRATS 2015 online testing set is given in Table \ref{Tab:2015_online_testing_compare} where the best and second best are highlighted. In this test set, our proposed network obtains competitive results compared against the state-of-the-art networks. In this experience, we also conduct an ablation to study the contribution of the detection network by reporting the results on BRATS 2018 validation set in two cases: with and without using the contextual detection network as shown in Table. \ref{tab:ablation}. 

\section{Conclusion}
In this paper, we have proposed a novel cascaded deep learning network approach for brain tumor detection and segmentation. Our network contains two targets, namely, brain tumor localization via our proposed contextual detection network and brain tumor segmentation via our proposed 3D atrous residual segmentation network. The proposed detector is able to effectively generate more meaningful proposals by only searching within the context-regions around ground-truth instances instead of entire an image pyramid. Our proposed network inherits the merits from both local and global information by making use of 3D atrous convolution with various kernel sizes without increasing the network size. Furthermore, it enables the gradient from the deep layers to be directly propagated to shallow layers through residual connections. The experimental results have proved that our proposed network outperforms other state of the art networks and can be considered to be a cutting-edge solution.

\bibliography{report.bib}
\bibliographystyle{ieeetr}

\end{document}